\documentclass[letterpaper, 10 pt, journal, twoside]{IEEEtran}
\usepackage{amsmath,amsfonts}
\usepackage{algorithmic}
\usepackage{algorithm}
\usepackage{array}
\usepackage[caption=false,font=normalsize,labelfont=sf,textfont=sf]{subfig}
\usepackage{textcomp}
\usepackage{stfloats}
\usepackage{url}
\usepackage{verbatim}
\usepackage{graphicx}
\usepackage{balance}
\usepackage{multirow, multicol}
\usepackage{makecell}
\usepackage{cite}
\usepackage[hang,flushmargin]{footmisc}
\usepackage{orcidlink}

\hypersetup{
    colorlinks,
    linkcolor={red!75},
    citecolor={blue!75},
    urlcolor={blue!75}
}

\RequirePackage{booktabs}

\title{Paper Title}

\author{First Author\\
Institution1\\
Institution1 address\\
{\tt\small firstauthor@i1.org}
\and
Second Author\\
Institution2\\
First line of institution2 address\\
{\tt\small secondauthor@i2.org}
}

\usepackage{enumitem}
\usepackage{multirow}
\usepackage{wrapfig}
\usepackage{colortbl}
\usepackage[normalem]{ulem}
\usepackage{xspace}

\setlist[itemize]{align=parleft,left=0pt}

\definecolor{kjs}{rgb}{0.36, 0.54, 0.66}
\definecolor{azure(colorwheel)}{rgb}{0.0, 0.5, 1.0}
\definecolor{nicegreen}{rgb}{0.0, 0.7, 0.1}
\definecolor{ashblue}{rgb}{0.36, 0.54, 0.66}
\definecolor{ashgrey}{rgb}{0.7, 0.75, 0.71}
\definecolor{applegreen}{rgb}{0.55, 0.71, 0.0}
\definecolor{jy}{rgb}{0.58, 0, 0.827}
\definecolor{cornellred}{rgb}{0.7, 0.11, 0.11}
\definecolor{darkcyan}{rgb}{0.0, 0.55, 0.55}
\definecolor{CuGray}{gray}{0.9}
\definecolor{airforceblue}{rgb}{0.36, 0.54, 0.66}
\definecolor{rev}{rgb}{0.784, 0.003, 0.313}
\definecolor{pink}{cmyk}{0, 0.7808, 0.4429, 0.1412}
\definecolor{amethyst}{rgb}{0.6, 0.4, 0.8}
\definecolor{black}{rgb}{0.0, 0.0, 0.0}
\definecolor{tb3_yellow}{rgb}{0.996, 1.0, 0.6}
\definecolor{tb3_orange}{rgb}{0.980, 0.8, 0.604}
\definecolor{tb3_red}{rgb}{0.972, 0.6, 0.6}
\definecolor{dimgray}{rgb}{0.41, 0.41, 0.41}
\definecolor{brickred}{rgb}{0.8, 0.25, 0.33}
\definecolor{bleudefrance}{rgb}{0.19, 0.55, 0.91}
\definecolor{blue(ncs)}{rgb}{0.265, 0.445, 0.765}
\definecolor{orange-red}{rgb}{1.0, 0.27, 0.0}

\newcolumntype{g}{>{\columncolor{CuGray}}c}
\newcolumntype{z}{>{\columncolor{CuGray}}l}

\renewcommand{\paragraph}[1]{\vspace{1mm}\noindent\textbf{#1.}\,\,}

\newcommand{\jskim}[1]{\textcolor{black}{#1}}

\usepackage{xspace}

\makeatletter
\def\@fnsymbol#1{\ensuremath{\ifcase#1\or *\or \dagger\or \ddagger\or
   \mathsection\or \mathparagraph\or \|\or **\or \dagger\dagger
   \or \ddagger\ddagger \else\@ctrerr\fi}}
\makeatother

\def\onedot{.\@\xspace}
\def\eg{\emph{e.g}\onedot} 
\def\ie{\emph{i.e}\onedot} 
 
\def\etc{\emph{etc}\onedot}

\newcommand{\Eref}[1]{Eq.~(\ref{#1})}
\newcommand{\Fref}[1]{Fig.~\ref{#1}}
\newcommand{\Tref}[1]{Table~\ref{#1}}

\newcommand{\bm}{{\mathbf{m}}}

\newcommand{\be}{\begin{eqnarray}}
\newcommand{\ee}{\end{eqnarray}}
\newcommand{\bee}{\begin{eqnarray*}}
\newcommand{\eee}{\end{eqnarray*}}

\newcommand{\matrixb}{\left[ \begin{array}}
\newcommand{\matrixe}{\end{array} \right]}

\definecolor{green(ncs)}{rgb}{0.0, 0.62, 0.42}

\newcommand{\hashgrid}{{Hashgrid}\xspace}
\newcommand{\tensorf}{{TensoRF}\xspace}
\newcommand{\kplanes}{{K-planes}\xspace}

\usepackage{amsmath,amsfonts,bm}

\def\eqref#1{equation~\ref{#1}}

\def\1{\bm{1}}

\def\vb{{\bm{b}}}

\def\vd{{\bm{d}}}

\def\vh{{\bm{h}}}

\def\vx{{\bm{x}}}

\def\mB{{\bm{B}}}

\def\mH{{\bm{H}}}

\def\mM{{\bm{M}}}

\def\mX{{\bm{X}}}

\DeclareMathAlphabet{\mathsfit}{\encodingdefault}{\sfdefault}{m}{sl}
\SetMathAlphabet{\mathsfit}{bold}{\encodingdefault}{\sfdefault}{bx}{n}

\def\sR{{\mathbb{R}}}

\newcommand*{\crosssymbol}{%
    \text{%
      \raise 1ex\hbox{%
        \rlap{\vrule height.2pt depth.2pt width .75ex}%
        \hbox to .75ex{\hss\vrule height .5ex depth 1ex\hss}%
      }%
    }%
}

\begin{document}

\title{Factorized Multi-Resolution HashGrid for Efficient Neural Radiance Fields: Execution on Edge-Devices}
\author{Kim Jun-Seong\textsuperscript{1$\ast$}\orcidlink{0000-0001-7570-6508} 
\quad Mingyu Kim\textsuperscript{2$\ast$}\orcidlink{0000-0001-5082-7223} 
\quad GeonU Kim\textsuperscript{3}\orcidlink{0009-0009-0224-0060} 
\quad Tae-Hyun Oh\textsuperscript{4$\crosssymbol$}\orcidlink{0000-0003-0468-1571}
\quad Jin-Hwa Kim\textsuperscript{5,6$\crosssymbol$}\orcidlink{0000-0002-0423-0415}
\thanks{This work was supported by Institute of Information \& communications Technology Planning \& Evaluation (IITP) grant (No.RS-2022-II220290 (2022-0-00290), Visual Intelligence for Space-Time Understanding and Generation based on Multi-layered Visual Common Sense; No.RS-2019-II191906, Artificial Intelligence Graduate School Program(POSTECH)) and National Research Foundation of Korea (NRF) grant (No. RS-2024-00358135, Corner Vision: Learning to Look Around the Corner through Multi-modal Signals) funded by the Korea government(MSIT). A portion of this work was carried out during an internship at NAVER AI Lab.}
\thanks{ ${}^1$Dept. of E.E. 
${}^3$Grad. School of AI of POSTECH. 
${}^2$Dept. of CS, The Univ. of British Columbia. 
${}^4$School of Computing Engineering, KAIST. 
${}^5$NAVER AI Lab, South Korea. 
${}^6$ AI Institute of Seoul Nat'l Univ.}
\thanks{${}^\ast$ Equal contribution. $\crosssymbol$ Corresponding authors.}
\thanks{© 2024 IEEE. Personal use of this material is permitted. Permission from IEEE must be obtained for all other uses, in any current or future media, including reprinting/republishing this material for advertising or promotional purposes, creating new collective works, for resale or redistribution to servers or lists, or reuse of any copyrighted component of this work in other works. Published in IEEE Robotics and Automation Letters, vol.~9, no.~11, Nov.~2024. DOI: 10.1109/LRA.2024.3460419.}
}

\markboth
{IEEE ROBOTICS AND AUTOMATION LETTERS. PREPRINT VERSION. ACCEPTED AUGUST, 2024}
{Jun-Seong \MakeLowercase{\textit{et al.}}: Fact-Hash}

\maketitle

\begin{abstract}
We introduce Fact-Hash, a novel parameter-encoding method for training on-device neural radiance fields.
Neural Radiance Fields (NeRF) have proven pivotal in 3D representations,
but their applications are limited due to 
large computational resources.
On-device training can open large application fields, 
providing strength in communication limitations, privacy concerns, and fast adaptation to a frequently changing scene.
However, challenges such as limited resources (GPU memory, storage, and power) impede their deployment.
To handle this, we introduce Fact-Hash, a novel parameter-encoding merging Tensor Factorization and Hash-encoding techniques.
This integration offers two benefits: the use of rich high-resolution features and the few-shot robustness. 
In Fact-Hash, we project 3D coordinates into multiple lower-dimensional forms (2D or 1D) before applying the hash function and then aggregate them into a single feature. 
Comparative evaluations against state-of-the-art methods demonstrate Fact-Hash's superior memory efficiency, preserving quality and rendering speed.
Fact-Hash saves memory usage by over one-third while maintaining the PSNR values compared to previous encoding methods.
The on-device experiment validates the superiority of Fact-Hash compared to alternative positional encoding methods in computational efficiency and energy consumption.
These findings highlight Fact-Hash as a promising solution to improve feature grid representation, address memory constraints, and improve quality in various applications.
Project page: \url{https://facthash.github.io/}
\end{abstract}    

\section{Introduction}
\label{sec:intro}

\IEEEPARstart{N}{eural} Radiance Fields (NeRF)~\cite{mildenhall2021nerf} serve as a prominent approach to the compact representation of 3D information in the context of the generation of 3D content in AR / VR domains~\cite{Deng2021FoVNeRFFN}, autonomous driving~\cite{tesla2022}, and telepresence~\cite{lombardi2021mixture}. 
Originally, NeRFs that learn 3D objects from image data through volume rendering naturally demand a high level of computation, such as cutting-edge GPUs. Specifically, it requires a significant amount of memory consumption or considerable computation time for each operation. Recent research suggests using distillation to reduce computing by making smaller models from pre-trained models~\cite{chen2022mobilenerf, Girish_2023_ICCV, li2023compressing}. However, creating smaller models can be tricky. Whenever new data is added, we need to re-train the original and smaller models, doubling the effort. In a practical service, discrepancies between the model given to users and the training model can pose management challenges for the service provider.

An alternative approach is using parameter-efficient NeRFs \cite{tang2022compressible, shin2023binary}. Having fewer parameters helps build a more stable system by lowering communication costs between the server and the edge-device. Additionally, NeRFs on edge devices provide quick adaptation to rapidly changing environments, like road driving scenarios, through fine-tuning. Despite these efforts, applying NeRFs on edge-device is less highlighted due to the following issues:\textbf{ (1) More strict memory constraints, (2) Limitations in storage space, (3) Fast inference/training speeds, (4) Robustness in a few shot scenarios}. Running NeRFs on edge devices requires the capability to work with limited GPU memory and storage capacity, as well as low power consumption. This is especially challenging in scenarios like drones and autonomous driving,
which can capture large-scale environments, costing over 10GB per scene~\cite{tancik2022blocknerf, Turki_2022_CVPR}.

Recent advances in parametric encoding methods in NeRF have  
mitigated some restrictions, such as
improving model expressiveness and accelerating learning and inference by embedding learnable features~\cite{sun2022improved, fridovich2022plenoxels}.
These methods still require a large memory footprint. 
In response, two primary approaches are proposed to address memory bottlenecks: Tensor Factorization~\cite{chen2022tensorf, fridovich2023k} and Hash-encoding~\cite{muller2022instant}. 
These methods provide efficient data compression and rendering for on-device NeRF applications.
However, they are accompanied by limitations such as excessive computational complexity in the case of Tensor Factorization and significant performance decline in few-shot scenarios with Hash-encoding.

To address these challenges and leverage the strengths of existing encoding methods, we introduce Fact-Hash,
harmoniously integrating the strengths of both existing encoding methods, 
Hash-encoding and Tensor Factorization families.
Fact-hash encoding involves two key stages: a \textit{Map} process that projects three-dimensional coordinates into multiple lower-dimensional coordinates and a subsequent \textit{Reduce}  process aggregating hash features derived from each mapped coordinate. 
This approach yields a condensed representation method that significantly cuts down the number of parameters without change in the learning pipeline.
Our proposed method exhibits 
robustness to few-shot cases.
While traditional collisions were distributed randomly throughout the spatial domain, our method limits them to two dimensions, promising more efficient learning.
While the previous methods faced instability, ours offers better performance and ensures stability in extreme conditions. 
This efficiency provides streamlined communication between servers and on-devices.

In summary, this paper introduces a novel and versatile Hash-encoding method that bridges the gap between existing encoding techniques. Our approach seeks to balance memory efficiency and scene coherence, ultimately enhancing the efficiency and quality of neural rendering. We comprehensively evaluate our method, highlighting its advantages across various scenarios, including few-shot cases in restricted GPU memory scenarios.
Our contribution is summarized as below:
\setlist[itemize]{align=parleft,left=0pt,topsep=1mm,itemsep=0mm,parsep=1mm}
\begin{itemize}
    \item [$\bullet$] We suggest novel compact encoding methods merging Hash-encoding and Tensor Factorization methods, considering the application of on-device execution. Through extensive experiments,
    we verify that our method operates robustly across diverse environments.
    \item [$\bullet$] Through real on-device experiments under limited power conditions, we demonstrate our model with significantly fewer parameters, resulting in faster rendering speed and maintaining novel-view reconstruction quality.
\end{itemize}
\section{Related work}
\label{sec:related_work}

\paragraph{Neural radiance fields}  
Neural Radiance Fields (NeRF)~\cite{mildenhall2021nerf} represent a prominent technique for synthesizing novel views by reconstructing high-fidelity 3D scenes. 
It
involves the optimization of coordinate-based Multi-Layer Perceptrons (MLPs) to estimate color and density values within the 3D scene, employing differentiable volume rendering. 
The efficacy
of NeRF has led to the introduction of several applications~\cite{synergnerf, martin2021nerf, jun2022hdr, fprf} for novel view synthesis tasks, demonstrating notable compactness and high-detail rendering quality by encapsulating 3D scenes in a few megabytes.
While compact,
its reliance on frequency encoding poses computational challenges, as this nonparametric positional encoding method places substantial demands on MLP capacities, resulting in computational bottlenecks~\cite{mildenhall2021nerf}.
This bottleneck becomes particularly pronounced in on-device scenarios for frequency-encoding-based NeRF models.

\paragraph{Parametric encodings}  
To reduce 
MLP dependency in NeRF representation and improve 
model efficiency, parametric encoding approaches have been introduced,  
integrating learnable parameters into the structured 3D spatial domain,
such as Dense grids~\cite{sun2022improved} and Octrees~\cite{fridovich2022plenoxels}. 
They demonstrate 
 enhancing training speed and  expressiveness of the representations compared to NeRF by spending 
a large memory footprint of gigabytes as a notable trad-off.

Later, 
alternative compact representations of
parametric encoding are explored:
the Tensor Decomposition~\cite{chen2022tensorf, fridovich2023k,han2023nrff}, and the Hash-Encoding~\cite{muller2022instant}.
These methods provide efficient data compression and fast rendering, aligning better with the requirements of on-device NeRF applications than the aforementioned ones.
Our 
method is also in this direction but specifically takes into account the conditions required for on-device deployment.
We combine the best of both worlds of 
factorization and hashing encoding, which also has favorable operating points under the restricted power 
of devices.

\paragraph{On-device aspect of NeRF}
Presently, investigations are underway to use NeRF in constrained on-device scenarios where the resource is extremely limited.
Existing studies~\cite{rt-nerf,instant3d, icarus} developed dedicated hardware chips for on-device NeRF.

On the other hand, 
there have been efforts to develop algorithmic ways for favorable NeRF modeling.
One prevalent approach is to ``bake''
\jskim{parametric encoding~\cite{chen2022mobilenerf, Girish_2023_ICCV, li2023compressing, reduced3dgs, shin2023binary}. }
These methods
typically involve two-stage processes: initially 
training a high-fidelity NeRF from voluminous data and subsequently tailoring it for mobile suitability through post-processing steps.
The post-processing stages incorporate existing parametric encoding methods through practices such as distillation 
\cite{cao2023real}, quantization~\cite{li2023compressing, Girish_2023_ICCV, chen2022mobilenerf}, and pruning~\cite{deng_wacv}.
Another direction involves adopting regularization to diminish model memory, such as rank priority~\cite{tang2022compressible} or binarization
\cite{shin2023binary}.

However, both approaches are predominantly focused on the rendering process.
In training aspect, both methods are unsuitable as they increase computational overhead in GPU memory by suggesting additional processes on existing methods, \eg, optimizing NeRF as the pre-process, which already introduces bottlenecks.
We introduce a mobile-training-friendly model not only for rendering but also for facilitating direct mobile on-device training in a single-stage manner, prioritizing minimal memory usage during both training and inference phases.

\begin{figure}[t]
    \centering
    \includegraphics[width=0.85\linewidth]{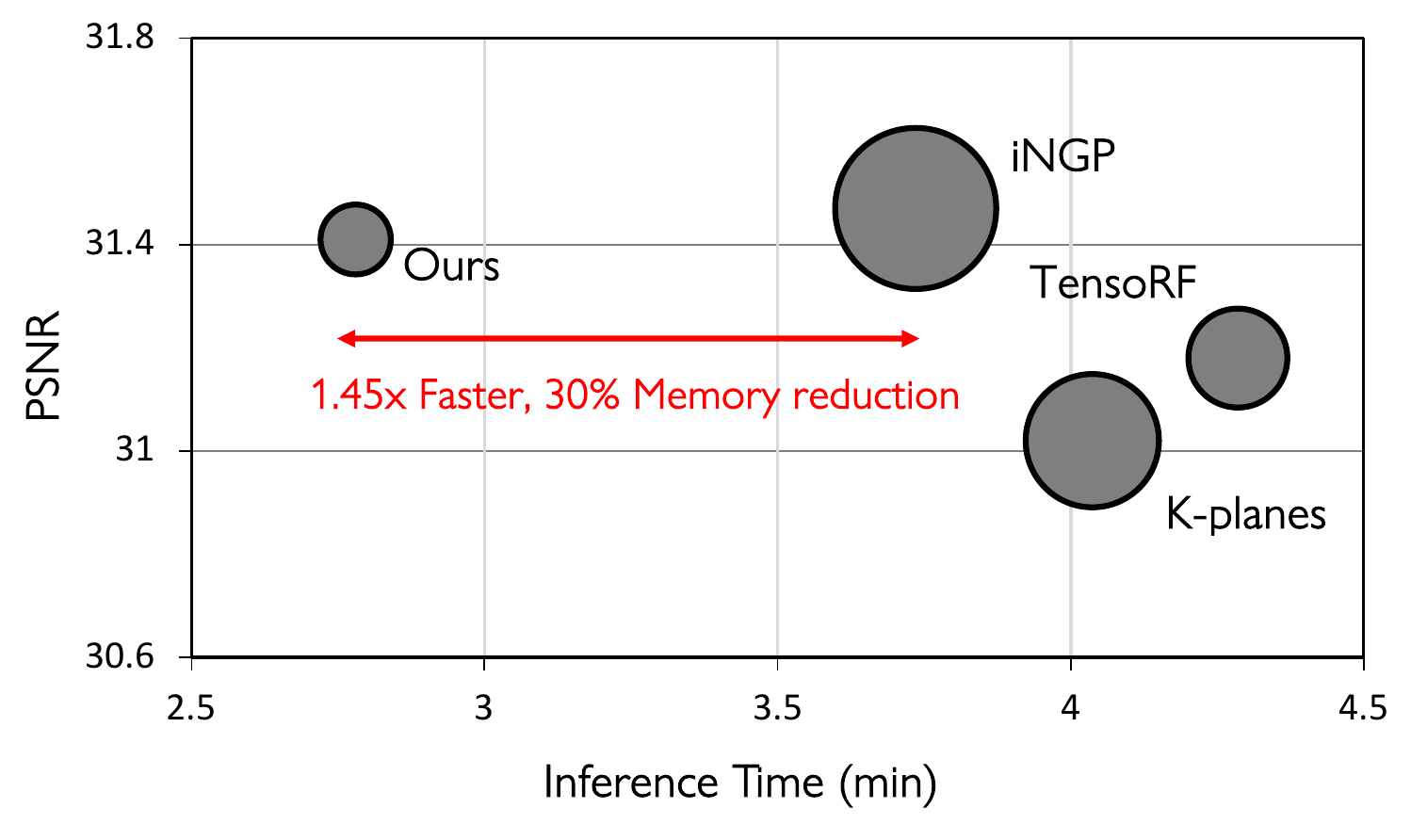}
    \vspace{-4mm}
    \caption{Comparison of Instant-ngp~\cite{muller2022instant}, TensoRF~\cite{chen2022tensorf}, K-planes~\cite{fridovich2023k} and Ours in terms of PSNR and inference time on the edge-device. Training is conducted on a standard GPU machine, whereas the inference is performed on the edge-device aligning with standard edge-device utilization practices. The area of each circle represents the model size.}
    \vspace{-2.5mm}
    \label{fig:full_view_qual}
    \vspace{-2.5mm}
\end{figure}
     
\section{Preliminary}
\label{sec:preliminary}

We aim to get the best of two worlds for positional encoding by using expressive multi-resolution and compact 
hashing, \hashgrid~\cite{muller2022instant}, and the Tensor Factorization methods such as \tensorf~\cite{chen2022tensorf} and \kplanes~\cite{fridovich2023k}.
Prior to an in-depth exploration of the suggested technique, an overview is provided regarding the foundational aspects of the neural rendering process, Multi-resolution Hashgrid, and Tensor Factorization.

\subsection{Neural rendering}
NeRF~\cite{mildenhall2021nerf} represents 3D volumetric scene with MLP$_\Phi$ that maps 3D coordinate $\vx=(x,y,z)$ and view-direction $\vd = (\phi,\theta)$ to a color $c$ and occupancy $\sigma$. 
Given a ray $\textbf{r}$ starting from the camera center $\textbf{o}$, through the direction $\textbf{d}$, we sample three dimensional coordinates $\textbf{p}_{k} = \textbf{o} + t_k \cdot \textbf{d}$,
where $k = 1,2,\cdots, K$. 
MLP$_\Phi$ predicts the color $c_{k}$ and occupancy $\sigma_{k}$ values for each point. 
Then, the color of the pixel corresponding to the ray $\textbf{r}$ is computed through the volume rendering.
Along the ray, we calculate the opacity value, the measure of how much the area is occupied, as:
\begin{align}
\label{eq:opacity}
    \alpha_k=1-\exp \left(-\sigma_k \delta_k\right), \delta_k=t_{k+1}-t_k .
\end{align}
Then, considering such opacity value, we calculate rendered color value $\hat{\textbf{C}}(\textbf{r})$ along the ray $\textbf{r}$ as,
\begin{align}
\label{eq:neural rendering}
    \hat{\textbf{C}}(\textbf{r})=\sum_{k=1}^{K} w_k \cdot {c}_k, \quad w_k=\prod_{l=1}^{k-1}\alpha_l \cdot \left(1-\alpha_l\right).
\end{align}

\begin{figure}[t]
    \centering
    \includegraphics[width=0.95\linewidth]{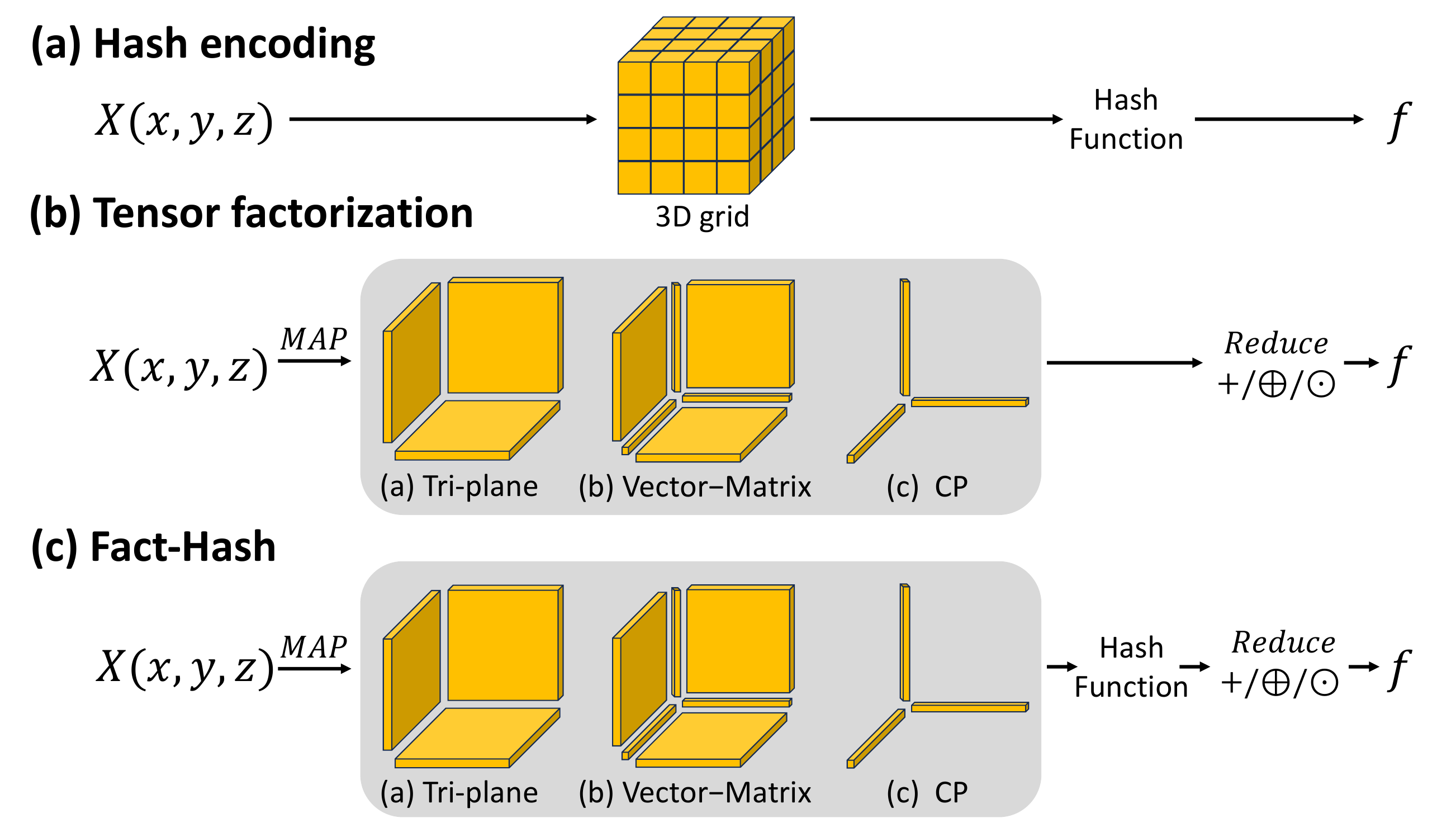}
    \vspace{-2.5mm}
    \caption{Conceptual illustration of parameter-encoding; iNGP~\cite{muller2022instant}, TensoRF~\cite{chen2022tensorf}, K-planes~\cite{fridovich2023k} and the proposed method. }
    \vspace{-2.5mm}
    \label{fig:pipeline}
    \vspace{-2.5mm}
\end{figure}

\subsection{Multi-resolution \hashgrid}
\label{sec:mulres_hashgrid}
Multi-resolution \hashgrid~\cite{muller2022instant} $f(\vx)$ processes a \jskim{$D$ dimensional coordinate,~\ie $D=3$ for 3D case,} $\vx \in [0,1]^{D}$ to an embedding vector $\vh \in \sR^{L\cdot F}$ parameterized by the $L$-level hash tables $\mH = \left \{ \mH_{l} \mid \mH_{l} \in \sR^{T \times F}, l \in \{1,2,\dots,L\} \right \}$.
Multi-resolution grids represent the levels at which each resolution (or the number of grids for each axis) is $N_{l}$ that varies exponentially according to 
the level $l$.
The number of grids increases exponentially from the coarsest $N_{\text{min}}$ to the finest $N_{\text{max}}$. 
Specifically, $N_l$ is defined as follows:
\begin{align}
\label{eq:hashgrid}
    N_{l} = 
    \lfloor N_{\text{min}} \cdot b^{l-1} \rfloor,   
    ~~\text{where}~~
    b = 
    \exp \left( 
        \tfrac{\ln N_{\text{max}} / N_{\text{min}}}{L-1}  
    \right). 
\end{align}

Under the hood, $f$ scales an input coordinate $\vx$ by $N_{l}$ to find the nearest $2^D$ grid vertices that form a unit hypercube using $\lfloor \vx \cdot N_{l} \rfloor$ and $\lceil \vx \cdot N_{l} \rceil$.
The corresponding entries in $\mH_l$ are recovered using a hash function $h(\vx)$ with the coordinates of the grid vertices, then reduced to a feature vector $\vh_l$ using trilinear interpolation weighing the distances to the grid vertices.
The output $\vh$ is the concatenation of the multi-resolution feature vectors $\left\{\vh_l \mid l \in \left\{1,2,\dots,L\right\}\right\}$. 
Remarkably, \cite{muller2022instant} shows that a small MLP is enough to outperform competitive NeRFs models. The sparse 3D grid structure of \hashgrid has a memory footprint log-linear to the resolution,
allowing versatile controllability using the growing factor $b$ and the size of hash table $T$. 
Exploiting a random hash encoding enables
to utilize the strength of higher resolution features (\Fref{fig:pipeline}-(a)).

\subsection{Tensor Factorization}
Tensor Factorization methods~\cite{chen2022tensorf, fridovich2023k, han2023nrff} represent the 3D grid structure using a combination of low-rank factors (\ie, planes and vectors).
These methods are based on a map-reduce pattern (see \Fref{fig:pipeline}-(b)).
First, the three-dimensional coordinates are \textit{Mapped} to lower-dimensional coordinate values.
For projected coordinates, linear and bilinear interpolations are performed for the vector and matrix, respectively, achieving low computation, equivalent to trilinear interpolation for a 3D tensor. 
The output of the per-axis features is then \textit{Reduced} to a single feature vector $\vh$ using element-wise operations, \eg, Hadamard product, concatenation, summation, or procedural combination of both.
The memory complexity is $\mathcal{O}(N^2)$, and further savings in memory footprint are possible.

Multiscale tensor decomposition~\cite{han2023nrff} and K-planes~\cite{fridovich2023k} demonstrate that extending Tensor Factorization to high-resolution multiscale features can further improve the representation power. 
However, 
Tensor Factorization 
still lacks scalability, requiring a capacity of the square of the resolution.

\section{Method}
\begin{figure*}[t]
    \centering
    \includegraphics[width=0.75\linewidth]{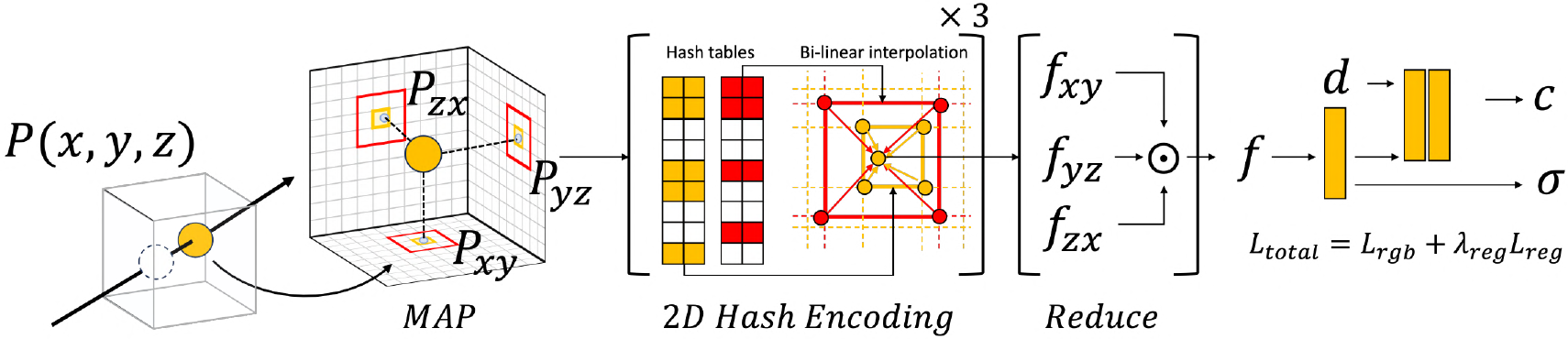}
    \vspace{-1.5mm}
    \caption{Schematic of the proposed method, Fact-Hash. For a given point $P$, we first project the point into the tri-plane and obtain surrounding multi-resolution 2D indices of each plane. For the multi-resolution 2D indices, we lookup the corresponding features from the assigned hash tables, and
    perform bi-linear interpolation to
    compose 3 multi-resolution feature vectors $f_{xy}, f_{yz},$ and $ f_{zx}$. 
    The interpolated feature vectors are multiplied with Hadamard product and decoded to the density $\sigma$ and the color $c$ with the shallow MLP decoders. We predict the color of ray using standard volumetric rendering formula with the densities and the colors along the ray and minimize the reconstruction loss with the regularization.   
    }
    \vspace{-4mm}
    \label{fig:overall_pipeline}
\end{figure*}
    
We propose Fact-Hash, a memory-efficient and fast rendering NeRFs method designed for edge-device services.
We first describe a generalized formula for integrating Hashgrid with Tensor Factorization.
Then, we introduce the Fact-Hash method, which provides a memory-efficient approach without sacrificing performance. We then proceed with the losses and the efficient rendering pipeline with rejection sampling to accelerate the rendering process further.

\subsection{Generalization}
We propose a generalization of Hashgrid inspired by Tensor Factorization that provides a unified view and unique opportunities to improve positional encoding.
First, we describe our proposed method and show how it can be instantiated as Hashgrid, Tensor Factorization methods, and even the multi-resolution mechanism using a map-reduce pattern. 

We generalize the Hashgrid by adopting the map-reduce pattern. 
Before applying Hashgrid $f(\vx)$, a tensor $\mM \in \sR^{D \times G \times D}$ \textit{maps} $\vx \in \sR^{D \times 1 \times 1}$ to $\mX' \in \sR^{G \times D}$ using the Hadamard product sequence using broadcast and sum in the first dimension, \jskim{where $D$ indicates input dimension, and $G$ indicates the number of mapped elements from each input coordinate.}
We may have a binary mask $\mB \in \sR^{G \times T}$, which limits the range of the hash function for each $G$ element.
Then, the $G$ outputs of $f(\vx')$ are \textit{reduced} to a single feature vector $\vh$ using elemental operations, \eg, Hadamard product, concatenation, or procedural combination of both.

\paragraph{Tensor Factorization method} 
We can also define vector-matrix decomposition using  above generalization. 
To formulate, a binary tensor $\mM \in \sR^{D \times 2D \times D}$ is defined as follows:
\begin{align}
    \mM[:,0,:] = \begin{bmatrix}
        1, 0, 0\\
        0, 1, 0\\
        0, 0, 0
    \end{bmatrix},
    \mM[:,1,:] = \begin{bmatrix}
        0, 0, 0\\
        0, 0, 0\\
        0, 0, 1
    \end{bmatrix}, \cdots
\end{align}
where $D=3$ for an illustrative purpose, without loss of generality. 
The first two mapped coordinates are $(x, y, 0)$ and $(0, 0, z)$. The other four are $(0, y, z)$, $(x, 0, 0)$, $(x, 0, z)$, and $(0, y, 0)$.
A proper choice of the binary vector $\vb$, a reducing function of the combination of element-wise `, and a sufficiently large $T$ make it equivalent to the vector-matrix decomposition of 
\tensorf. 
Recall that \hashgrid attempts one-to-one mapping under sufficiently large $T$ of a hash table~\cite{muller2022instant}.
Similarly, by removing the vector projection from the binary mask, we can also derive a triplane decomposition~\cite{fridovich2023k}.

\paragraph{Multi-resolution} A 
binary tensor
$\mM \in \sR^{D \times L \times D}$ is used for mapping where the diagonal elements of $\left\{\mM[:,l,:] \mid l \in \left\{1, 2, \dots, L\right\}\right\}$ are $N_1, N_2, \dots, N_L$, respectively. Again, a proper binary mask allocates the entries in a hash table equivalently to the entries of individual hash tables dedicated to each level. 

\subsection{Fact-Hash}
Now, we
introduce
how Fact-Hash efficiently parameterizes NeRFs.
In terms of generalized formula, Fact-Hash utilizes tri-plane decomposition as a \textit{map} function and Hadamard multiplication as a \textit{reduce} function.
As depicted in \Fref{fig:pipeline}-(a,b), both methods exhibit different compression strategies. 
In Fact-Hash, we subsequently
operate a hash function for the factorized features. \Fref{fig:pipeline}-(c) illustrates the overall pipeline of Fact-Hash.
Employing separate hash tables for each axis in our method
aims to minimize collisions within two dimensions, diverging from widespread collisions prevalent across the entire three-dimensional space.
The overall pipeline is illustrated in \Fref{fig:overall_pipeline}.

\paragraph{Multiscale features} We utilize multiscale features to find a compressive representation while maintaining expressive power. Empirical evidence from prior research indicates that multiscale plane features exhibit higher expressive capabilities compared to single-resolution~\cite{muller2022instant, fridovich2023k, han2023nrff}. This observation underscores the potential
of achieving similar expressive power with fewer parameters. Several Tensor Factorization methodologies explore multiscale expansion to harness high-resolution features~\cite{han2023nrff, fridovich2023k}. Leveraging this multi-resolution grid approach, each plane feature in our model incorporates a multi-resolution grid.
As Hash-encoding thresholds the upper limit of memory consumption, our model can effectively harness multiscale attributes without a significant surge in storage space thanks to
the 
integration of the hash function.

\paragraph{Hybrid modeling} We integrate the expressive capabilities of MLP
with the parametric encoding method 
to explore an operating regime favorable to the on-device scenario with
the hybrid model's performance and expressive capacity.

\subsection{Inference}
In this context, we describe the procedures to extract features from given 3D point queries. 
First, the three-dimensional point $P$ undergoes the Map process, where $P$ is projected to two-dimensional points $P_{xy}$, $P_{yz}$, and $P_{zx}$ on the tri-planes. 
For each two-dimensional point, the nearest four coordinates are sampled and Hash-encoded using distinct hash tables. 
Each hash feature is bilinearly interpolated to obtain plane features $f_{xy}$, $f_{yz}$, and $f_{zx}$, respectively. 
Then, Reduce process is executed on obtained features. 
Fact-Hash uses the Hadamard product to utilize its compactness in intermediate features.

The final feature f is fed to the MLP decoders with the viewing direction. 
We use color and occupancy networks. 
The occupancy is determined from the first channel of the output of the first network. 
The viewing direction is encoded by the spherical harmonic function of degree four. 
The process repeats through the samples along the ray. 
Then, the 
subsequent 
volume-rendering process 
regresses the pixel values of the ray.

\subsection{Losses}
We describe the loss terms we use to optimize our model.
First, we calculate the photometric loss $\mathcal{L}_\textit{rgb}$. 
Photometric loss is calculated by mean square errors between the rendered pixel values $\hat{\textbf{C}}(\textbf{r})$, and the ground truth pixel values $\textbf{C}_{gt}(\textbf{r})$,
\begin{align}
\label{eq:photo_loss}
    \mathcal{L}_\textit{rgb} = \sum\nolimits_{r\in R} \lVert \hat{\textbf{C}}(\textbf{r}) - \textbf{C}_{gt}(\textbf{r}) \rVert^2.
\end{align}
Rendered pixel values $\hat{\textbf{C}}(\textbf{r})$ are calculated by ray-marching of samples along the ray as \Eref{eq:neural rendering}.

To promote the model to predict whether a point is
occupied or vacant and to reduce vague floating artifacts, we add the opacity regularization term. We add the entropy loss term on the opacity value to encourage such an effect.
The opacity $\alpha$ is calculated as \Eref{eq:opacity}.
\begin{align}
\label{eq:opacity_loss}
    \mathcal{L}_\textit{op} = \sum\nolimits_{k\in K}-\alpha_k \cdot \log(\alpha_k).
\end{align}

To further reduce floating artifacts, the model integrates the distortion loss $\mathcal{L}_{dist}$ suggested in
\cite{barron2022mipnerf360}. 
Distortion loss promotes gathering 
volume rendering weights and mitigates issues such as floating artifacts and background collapses.
For computational efficiency, we adopt the efficient implementation of the distortion loss
~\cite{sun2022improved}.
\jskim{The overall loss term is summarized as follows: 
\begin{align}
\label{eq:loss_total}
    \mathcal{L}_\textit{total} = \mathcal{L}_\textit{rgb} + \lambda_{op} \mathcal{L}_\textit{op}  + \lambda_{dist} \mathcal{L}_\textit{dist},
\end{align}
where we set 
the weights of $\lambda_{op} = 1e^{-3}$ for the occupancy loss and $\lambda_{dist} = 1e^{-2}$ for the distortion loss.}

\subsection{Sample rejection for efficient rendering}
We exploit 
a rejection sampling method
to reduce the computational workload by allowing to filter out samples in empty spaces.
TensoRF\cite{chen2022tensorf}, K-planes\cite{fridovich2023k}, and Instant-NGP\cite{muller2022instant} 
encompass different rejection sampling techniques: alpha-grid, proposal network, and density bit-field respectively. 
In scenes with low occupancy, the bitfield method shows potential for lower computation and faster speed, making it more suitable for 
on-device situations.
The proposal network may face speed bottlenecks due to computational constraints, but experimental evidence confirms its robust operation in larger scenes with a coarse-to-fine strategy.
Therefore, we selectively consider these two methods on the basis of the scene's characteristics.
For further reduction, sampling termination method is also applied, which avoids additional sampling for rays 
when the ray transmittance falls below a threshold.

\section{Experiment}
\label{sec:experiments}
In this section, we present our experiments designed to address three pivotal questions: \textit{1) Do existing fast neural radiance field techniques offer a cost-effective way for server and edge-devices to communicate, considering the size of the parameters involved?} \textit{2) Do fast radiance fields deliver good performance when there's limited storage space?} \textit{3) Do NeRFs support fast execution on edge-device machines?} 

To answer those questions, we carried out vast experiments with both synthetic and real-world dataset. As baselines, we employ cutting-edge parametric encoding methodologies, denoted as state-of-the-art (SOTA) techniques, namely iNGP~\cite{muller2022instant}, TensoRF~\cite{chen2022tensorf}, and K-planes~\cite{fridovich2023k}. The aim was to assess the suitability of these approaches in edge-device scenarios with limited GPU memory. 

Moreover, we also include an ablation study to provide evidence for the feasibility of running NeRFs on edge devices. In this experiment, we selected the Jetson Xavier NX 16GB~\cite{jetsonnx} as our reference machine. This choice was made due to its electric power control unit and CUDA compatibility, allowing us to run baselines without additional effort. In other aspect, an algorithm exceeding this memory limit was excluded, as it wouldn't be suitable for standard edge devices. To address this constraint, we introduce re-implementations of TensoRF and K-planes. These versions are designed to meet the memory requirement and fast rendering speed, incorporating efficient sampling rejection and in-house CUDA implementation. We name these versions as \texttt{TensoRF\crosssymbol} and \texttt{K-planes\crosssymbol}.

\begin{table}[t]
    \caption{Experimental results on the NeRF synthetic datasets.
    }
    \label{table:nerf_synthetic}
    \centering
    \resizebox{0.9\linewidth}{!}{%
\begin{tabular}{c|lcc|ccc}
\toprule
\multirow{2}{*}{Case}                                                  & \multicolumn{3}{c|}{Model} & \multirow{2}{*}{\begin{tabular}[c]{@{}c@{}}Avg. \\ PSNR$\uparrow$\end{tabular}} & \multirow{2}{*}{\begin{tabular}[c]{@{}c@{}}Avg. \\ SSIM$\uparrow$\end{tabular}} & \multirow{2}{*}{\begin{tabular}[c]{@{}c@{}}Avg. \\ LPIPS$\downarrow$\end{tabular}} \\
                                                                       \cmidrule{2-4} 
                                                                       & \multicolumn{1}{l}{Name}  &    \#Params (M) $\downarrow$ & Size$ \downarrow$~(Mb)        &                                                                       &                                                                        \\
\midrule
\multirow{6}{*}{\begin{tabular}[c]{@{}c@{}}Full \\ views\end{tabular}} & iNGP                   &   11.46   &   45.24   &   31.59   &   0.953   &   0.061   \\
                                                                       & TensoRF                &   17.38   &   68.69   &   32.88   &   0.961   &   0.028   \\
                                                                       & K-planes               &   33.04   &   393.08  &   32.36   &   0.962   &   0.049   \\
                                                                       \cmidrule{2-7}
                                                                       & TensoRF\crosssymbol   &   6.51    &   25.73   &   31.69   &   0.951   &   0.064   \\
                                                                       & K-planes\crosssymbol   &   16.53   &   64.82   &   32.06   &   0.956   &   0.057   \\
                                                                       & Ours                   &   3.40    &   13.54   &   32.00   &   0.954   &   0.060   \\
                                                                        \midrule
\multirow{6}{*}{8 views}                                               & iNGP                   &   11.46   &   45.24   &   18.38   &   0.811   &   0.246   \\
                                                                       & TensoRF                &   17.38   &   17.11   &   22.78   &   0.854   &   0.142   \\
                                                                       & K-planes               &   33.04   &   393.08  &   24.41   &   0.899   &   0.106   \\
                                                                       \cmidrule{2-7}
                                                                       & TensoRF\crosssymbol   &   6.51    &   25.73   &   22.98   &   0.860   &   0.158   \\
                                                                       & K-planes\crosssymbol   &   16.53   &   64.82   &   23.58   &   0.882   &   0.127   \\
                                                                       & Ours                   &   3.40    &   13.54   &   23.68   &   0.877   &   0.129   \\
                                                                       \bottomrule
\end{tabular}
}%
\vspace{-2.5mm}
\end{table}

Throughout extensive experiments, the design efficacy of our model is validated in three key aspects: the number of parameters, comparable performance and operation time. We illustrate results on synthetic datasets, rendering speed on the edge-device, and real-world applications in sequence.

\subsection{Synthetic dataset: NeRF Synthetic}
\label{sec:codebase_comparison}
\begin{figure}[!t]
    \centering
    \includegraphics[width=0.9\linewidth]{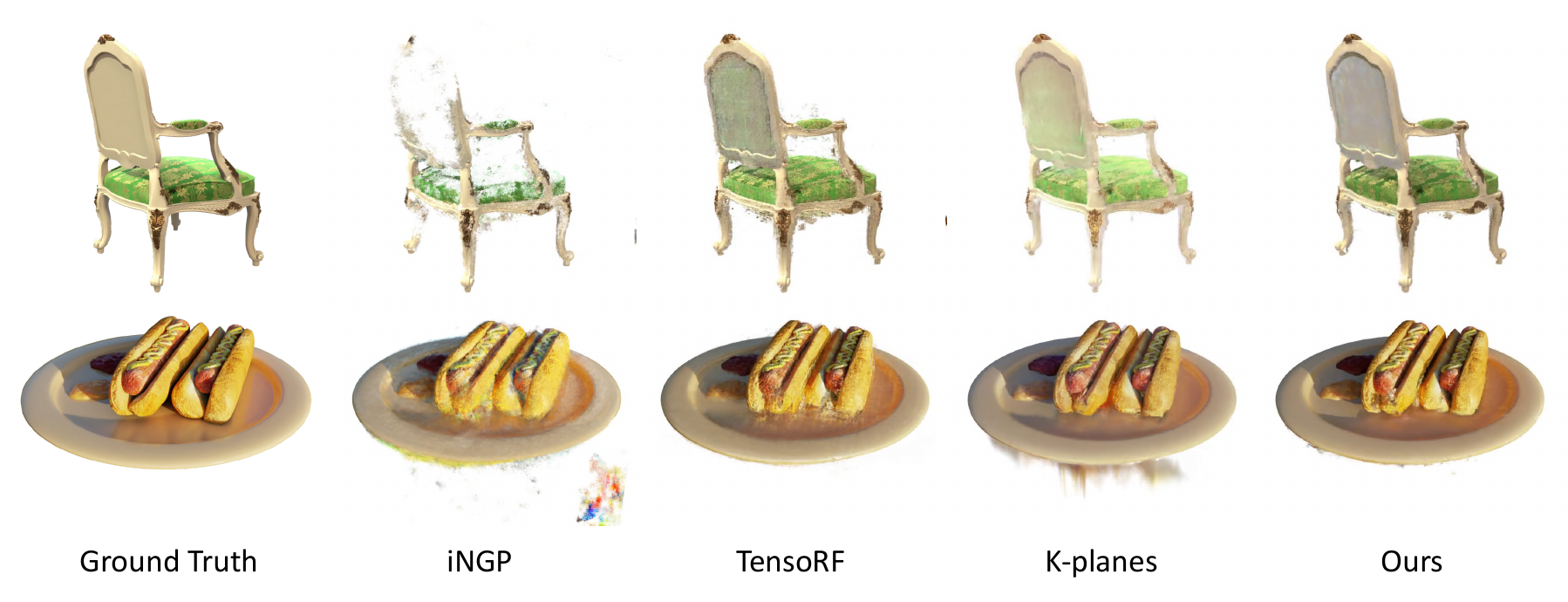}\vspace{-2mm}
    \caption{Qualitative results of 8 views case on the NeRF synthetic dataset. Rendered images are results of \textit{chair}, \textit{hotdog} cases in the NeRF synthetic dataset by iNGP, TensoRF, K-planes, and ours.}
    \vspace{-2.5mm}
    \label{fig:few_shot_qual}
\end{figure}
\begin{figure}[!t]
    \centering
    \includegraphics[width=0.9\linewidth]{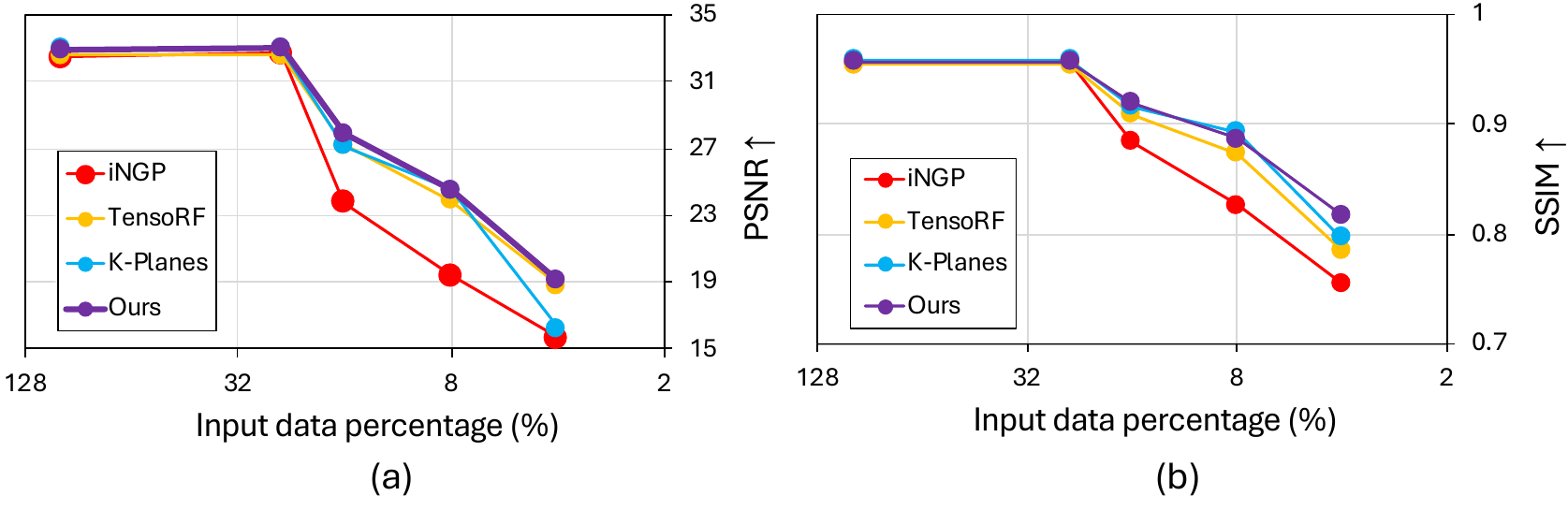}\vspace{-2mm}
    \caption{\jskim{PSNR (a) and SSIM (b) values according to the number of uniformly sampled inputs. All metrics are average value of 7 NeRF synthetic data.}}
    \vspace{-2.5mm}
    \label{fig:various_few_shot}
    \vspace{-2.5mm}
\end{figure}
In this subsection, we aim to demonstrate that our proposed method achieves comparable performance to the baselines while requiring an extremely low number of parameters under normal conditions. To emphasize the performance comparison, we conduct both training and evaluation on an RTX3090Ti.

The NeRF synthetic dataset consists of images rendered for eight objects within a confined area, comprising 100 training images and 200 test images. To address various real-world applications, we employed two training strategies. One follows the standard NeRF Synthetic training scheme, whereas the other involves a sparse-inputs experiment. 
\jskim{For the sparse-inputs, we uniformly sample views following prior works\cite{yang2023freenerf, jain2021putting}}.
We evaluated using full-resolution images (800 $\times$ 800 pixels) for both training and testing, and use PSNR, SSIM, and LPIPS as evaluation metrics. Our baselines included not only iNGP, TensoRF, and K-planes but also re-implementations of TensoRF and K-planes, designed for on-device execution.

\jskim{For fair comparison, all experiments were conducted with the batch 4096, and iteration 30K. 
We minimally adjusted hyperparameters to fit the on-device constraints while maximizing the performance.
For Baseline, we modify TensoRF initial grid resolution from $128^3$ to $32^3$. 
For re-implementation, we reduces \texttt{TensoRF\crosssymbol} channel from 64 to 24, and \texttt{K-planes\crosssymbol} channel from 32 to 16. 
}

The experimental results are presented in \Tref{table:nerf_synthetic}. 
Initially, our proposed method showcases a significant reduction in the number of parameters from 1/2 to 1/5 when compared to the baselines. Despite the reduction, our method achieves competitive or superior performance compared to baselines. Specifically, while the average PSNR scores are ranked as TensoRF, K-planes, Ours, and iNGP, it's important to note that the differences in performance are not substantially remarkable. 

However, in the few-shot scenario, iNGP experiences a significant decline in performance compared to the standard training strategy. In contrast, TensoRF and K-planes, which leverage multi-plane representation, demonstrate greater resilience to overfitting issues (see \Fref{fig:few_shot_qual}), consistent with previous observation~\cite{synergnerf}. Notably, K-planes, incorporating TV loss, exhibit smoothing effects, while TensoRF exhibits artifacts resembling scratches. 
Although our approach does exhibit some artifacts, it's clear that we achieve more favorable results compared to alternative models. 
\jskim{To confirm diverse cases, we conducted experiments on four sparse cases, 4\%, 8\%, 16\%, and 24\% (see~\Fref{fig:various_few_shot}). Our model consistently demonstrate robust performance in terms of  PSNR and SSIM compared to other models.}
Notably, in sparse-input scenarios, our method outperforms alternatives, including \texttt{K-planes}\crosssymbol and \texttt{TensoRF}\crosssymbol, underscoring the robustness of proposed model against overfitting.

The rendered results underscore the superiority of our model, as shown in \Fref{fig:full_view_qual}, particularly evident in the high-resolution feature, facilitating enhanced learning about intricate structures. Examination of micro-mesh details reveals artifacts in the case of K-planes or TensoRF when employing lower-resolution expressions. Instant-ngp, while occasionally displaying noisy rendering results, is a known issue associated with independently updating resolution features, consistent with observations in other research~\cite{wu2023neural}. Nonetheless, rendering in the factorized form often mitigates such artifacts. \jskim{For 360 degree view rendering please refer supplementary videos.}

\begin{figure}[!t]
    \centering
    \includegraphics[width=0.9\linewidth]{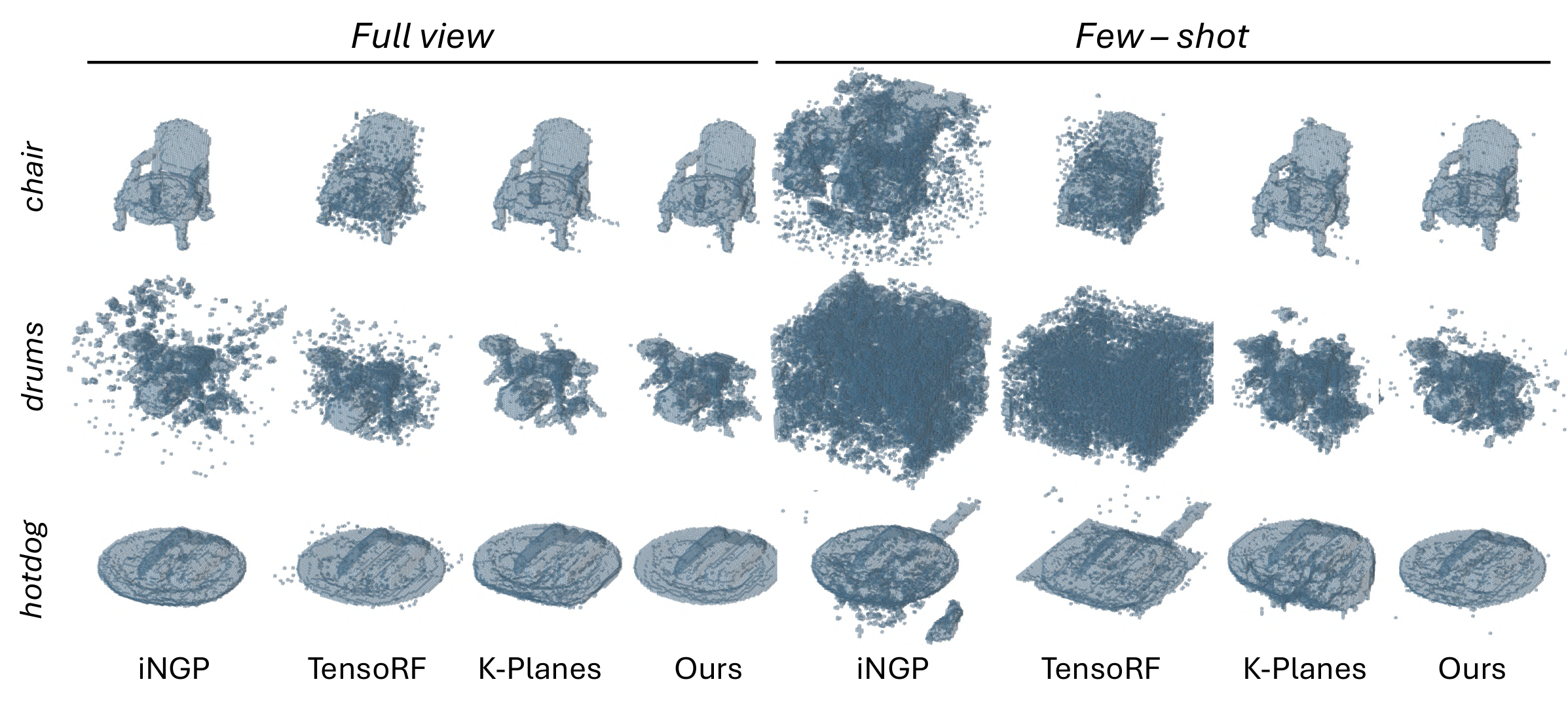}\vspace{-2mm}
    \caption{Visualization of bitfield on the NeRF synthetic dataset. Each case is trained on 8-views on \textit{chair}, \textit{drums} and \textit{hotdog}.}\vspace{-4mm}
    \label{fig:bitfield}
    \vspace{-2.5mm}
\end{figure}
The superiority of our model is better demonstrated in the learned bitfield results.
\Fref{fig:bitfield} shows the visualization of the bitfield learned in each model directly.
In controlled setting, other methods show more inaccurate overfitted bitfield results occurring floating artifacts.
In the case of voxel grid-based iNGP, such floating artifacts especially severe showing tendency to overfit without accurately learning the location of the object.
Using plane feature leads to learn bitfield more tightly to objects, but TensoRF and K-planes, which lack expressive power, results in inaccurate understanding in occupancy.
In comparison, our model learned object-tight opacity in most scenes, which demonstrate the robustness of our model, while maintaining high compactness and expressiveness.

\subsection{Rendering time on edge-device}
\label{sec:on_device}
\begin{table}[!t]
    \caption{rendering time for 200 images on the NeRF synthetic datasets.
    } 
    \label{table:edge_device}
    \centering
    \resizebox{1.0\linewidth}{!}{%
\begin{tabular}{l|cccccccc|c|c}
\toprule
\multicolumn{1}{c|}{\multirow{2}{*}{Model}} & \multicolumn{9}{c|}{10 Watt $\downarrow$}                                                    & \multirow{2}{*}{\begin{tabular}[c]{@{}c@{}}20 Watt \\ Avg.$\downarrow$ \end{tabular}} \\
\cmidrule{2-10}
\multicolumn{1}{c|}{}                        & chair & drums & ficus & hotdog & lego  & materials & mic   & ship  & Avg.  &                                                                      \\
\midrule
iNGP                                        & 71.8  & 142.3 & 507.0 & 167.4  & 126.3 & 251.3     & 96.9  & 566.6 & 241.2 & 191.4                                                                \\
TensoRF\crosssymbol                                & 109.5 & 199.8 & 174.3 & 199.2  & 188.4 & 395.6     & 149.0 & 641.1 & 257.1 & 219.6                                                                \\
K-Planes\crosssymbol                              & 97.8  & 195.8 & 159.1 & 206.4  & 167.8 & 339.3     & 122.1 & 504.7 & 224.1 & 185.9                                                                \\
Ours                                        & 93.7  & 164.5 & 145.1 & 130.7  & 161.8 & 319.1     & 89.6  & 229.8 & 166.8 & 133.2                                                               \\
\bottomrule
\end{tabular}
\vspace{-4mm}
}%
\end{table}

We conducted experiments assuming real-world edge-device scenarios, wherein NeRFs models trained on GPU machines were utilized for novel-view synthesis rendering on edge devices. The primary objective of these experiments was to demonstrate that our proposed method offers faster rendering speeds on average compared to other baselines. One important point to mention is that we excluded the original K-planes and TensoRF methods and instead used the re-implemented versions, specifically \texttt{K-Plane\crosssymbol} and \texttt{TensoRF\crosssymbol}, enhanced with the efficient rendering technique used in iNGP\cite{muller2022instant}. Given that the original methods had longer rendering times, our goal was to compare them with our proposed method, aligning with an identical rendering pipeline. This approach leads to emphasize the efficiency of our proposed method under a fair comparison in terms of rendering speed.

A quantitative comparison is provided in \Tref{table:edge_device}. 
Firstly, the proposed method consistently exhibits faster rendering speeds at both 10-watt and 20-watt power settings. Specifically, iNGP, which uses a 3D Feature Grid, shows faster rendering speeds except for the \textit{ficus} and \textit{ship} scenes. Although the proposed method and K-planes perform three interpolations, and TensoRF requires six interpolations, iNGP uses only one interpolation, allowing quick rendering. However, in highly complex scenes like \textit{ficus} and \textit{ship}, iNGP exhibits significantly slower speeds than the proposed models. This is because, in extremely thin and complex scenes, iNGP requires more samples to maintain performance (see \Fref{fig:bitfield}). However, the proposed method, with its projection to a two-dimensional feature grid, which is dimensionally reduced compared to the 3D voxel grid, is less sensitive in specific cases, requiring fewer samples to achieve comparable performance.

Especially, by increasing rendering speed approximately 40\% on average compared to the existing methods, the proposed method can significantly enhance resource utilization and user convenience in applications that repeatedly perform NeRFs services on edge devices.
Note that in~\Tref{table:edge_device}, Fact-Hash utilized 20\% to 50\% params.~of the others.
Also, Fact-Hash is orthogonal with post-processing (quantization, pruning,~\etc) methods, which may further improve speed.

\subsection{Applicability in real-world scenarios}
\label{sec:Application}
\begin{table}[t]
    \caption{Experimental results on the tank and temples datasets.
    }
    \label{table:tank_and_temples}
    \centering
    \resizebox{0.9\linewidth}{!}{%
\begin{tabular}{c|lcc|ccc}
\toprule
\multirow{2}{*}{Case}                                                  & \multicolumn{3}{c|}{Model} & \multirow{2}{*}{\begin{tabular}[c]{@{}c@{}}Avg. \\ PSNR$\uparrow$\end{tabular}} & \multirow{2}{*}{\begin{tabular}[c]{@{}c@{}}Avg. \\ SSIM$\uparrow$\end{tabular}} & \multirow{2}{*}{\begin{tabular}[c]{@{}c@{}}Avg. \\ LPIPS$\downarrow$\end{tabular}} \\
\cmidrule{2-4} 
& \multicolumn{1}{l}{Name} & \#Params (M) $\downarrow$ & Size$ \downarrow$~(Mb) & & \\
\midrule
\multirow{4}{*}{\begin{tabular}[c]{@{}c@{}}Full \\ views\end{tabular}} & iNGP&   11.46   &   44.18   &   28.05   &   0.918   &   0.128   \\
 & TensoRF\crosssymbol   &   4.89    &   18.92   &   27.93   &   0.912   &   0.139   \\
 & K-planes\crosssymbol   &   16.53   &   63.30   &   28.40   &   0.923   &   0.120   \\ 
 & Ours   &   3.40    &   13.22   &   28.42   &   0.919   &   0.121  \\
 \midrule
\multirow{4}{*}{10\% views} & iNGP &   11.46   &   44.18   &   19.73   &   0.869   &   0.186   \\
 & TensoRF\crosssymbol   &   4.89    &   18.92   &   21.57   &   0.861   &   0.190   \\
 & K-planes\crosssymbol   &   16.53   &   63.30   &   22.28   &   0.878   &   0.152   \\
 & Ours &   3.40    &   13.22   &   22.69   &   0.876   &   0.151   \\
 \bottomrule
\end{tabular}
\vspace{-3.5mm}
}%
\end{table}
\begin{figure}[ht]
    \centering
    \includegraphics[width=0.9\linewidth]{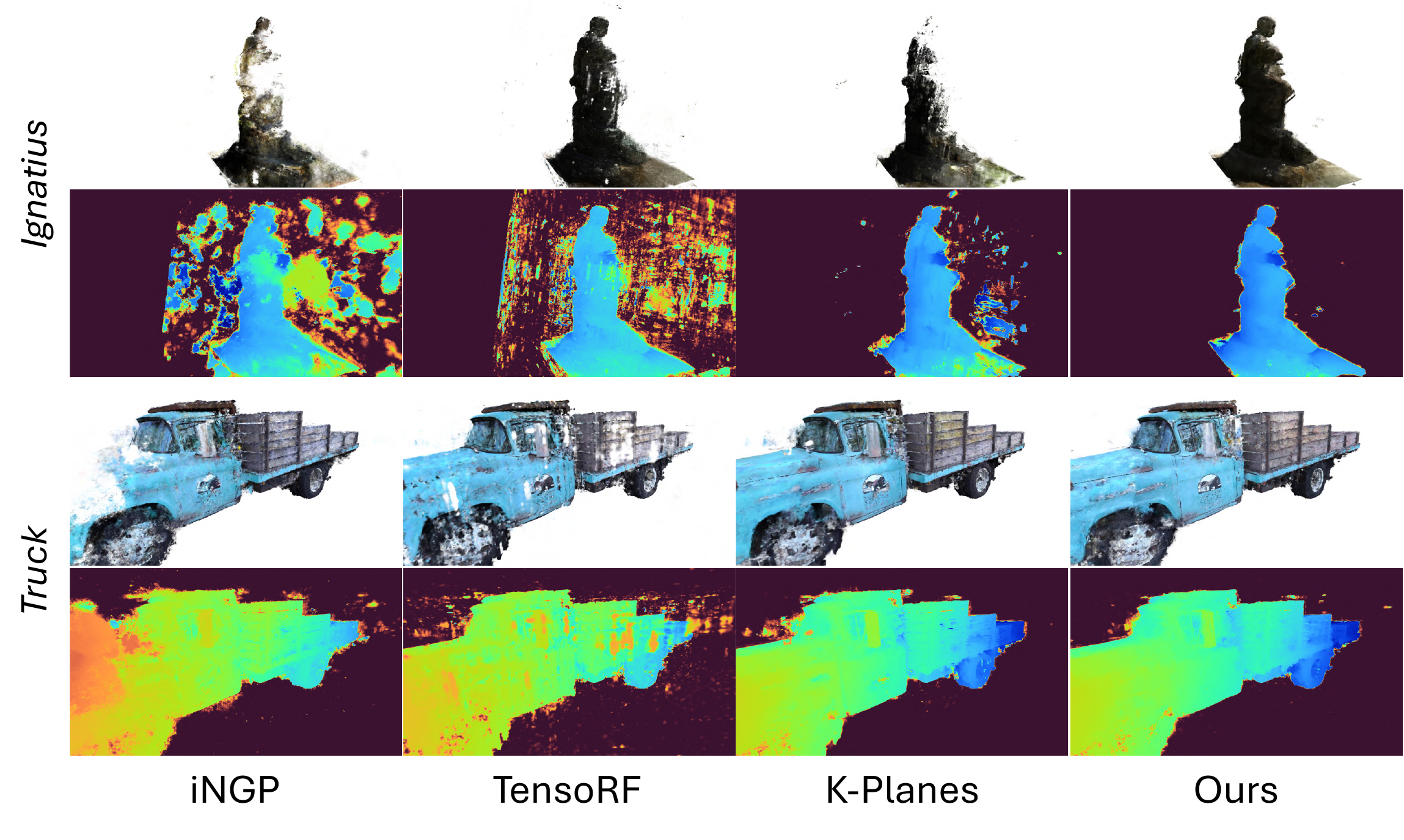}\vspace{-2mm}
    \caption{Qualitative results on the Tank and Temples for 10\% training inputs.}\vspace{-4mm}
    \vspace{-2.5mm}
    \label{fig:tnt_qual}
\end{figure}
The proposed model is highly compatible with large-scale scene reconstruction on edge-device, particularly regarding few-shot robustness and model size. Training a large-scale scene on the edge device can be limited by storage capacity, making it challenging to use a full-view training setting that involves uploading the entire dataset. 
The model size is also strictly constrained due to the necessity of saving multiple block-decomposed models~\cite{tancik2022blocknerf} for large-scale scene reconstruction. 
To extend our evaluation to a broader spatial context, we experiment on two real-world outdoor scenarios: high resolution 360$^\circ$ data, and block-scale data.

The Tank and Temples dataset is high resolution which have more than four times higher pixel resolutions than the synthetic NeRF dataset, and contain imperfections in lighting conditions, masking, etc.
We use four scenes (\textit{caterpillar, family, ignatius, truck}) for experiment.
We further experimented using scarse input setting by uniformly sampling 10\% of the inputs images.
Similar to \ref{sec:codebase_comparison}, we re-implemented baselines by grid-searching parameters to fit on-device memory constraint.

Numerical comparison demonstrates superior performance compared to baseline metrics across most metrics, as shown in \Tref{table:tank_and_temples}.
Experimental results demonstrate that our model exhibits robust learning capabilities in both scenarios where the input image is sufficient or few, akin to prior observations.
This suggests that the efficacy of the preceding experiment's results is not confined to specific instances but extends to general circumstances.
Moreover, in the comparison of rendering outcomes, the model's learning stability emerges prominently (see \Fref{fig:tnt_qual}).
Analysis of depth rendering results reveals that the proposed model can predict occupancy tightly to the objects, mitigating the occurrence of unnecessary floating artifacts, which cause white cloudy artifacts. \jskim{For 360 degree view rendering please refer supplementary videos.}

To further extend the evaluation to practical scenarios, we also conducted block-scale experiment in full-view and few-shot cases.
Reconstruction of block-scale scene requires us to store a model decomposed into many blocks~\cite{tancik2022blocknerf}, which severely limits the model size.
We use San Francisco Mission Bay dataset~\cite{tancik2022blocknerf}, an urban scene dataset consisting of approximately 12,000 images captured by 12 cameras which covers block-scale of areas.
We design a few-shot setting by sampling intervening images from each camera view with interval 3.

\begin{figure}[!t]
    \centering
    \includegraphics[width=0.9\linewidth]{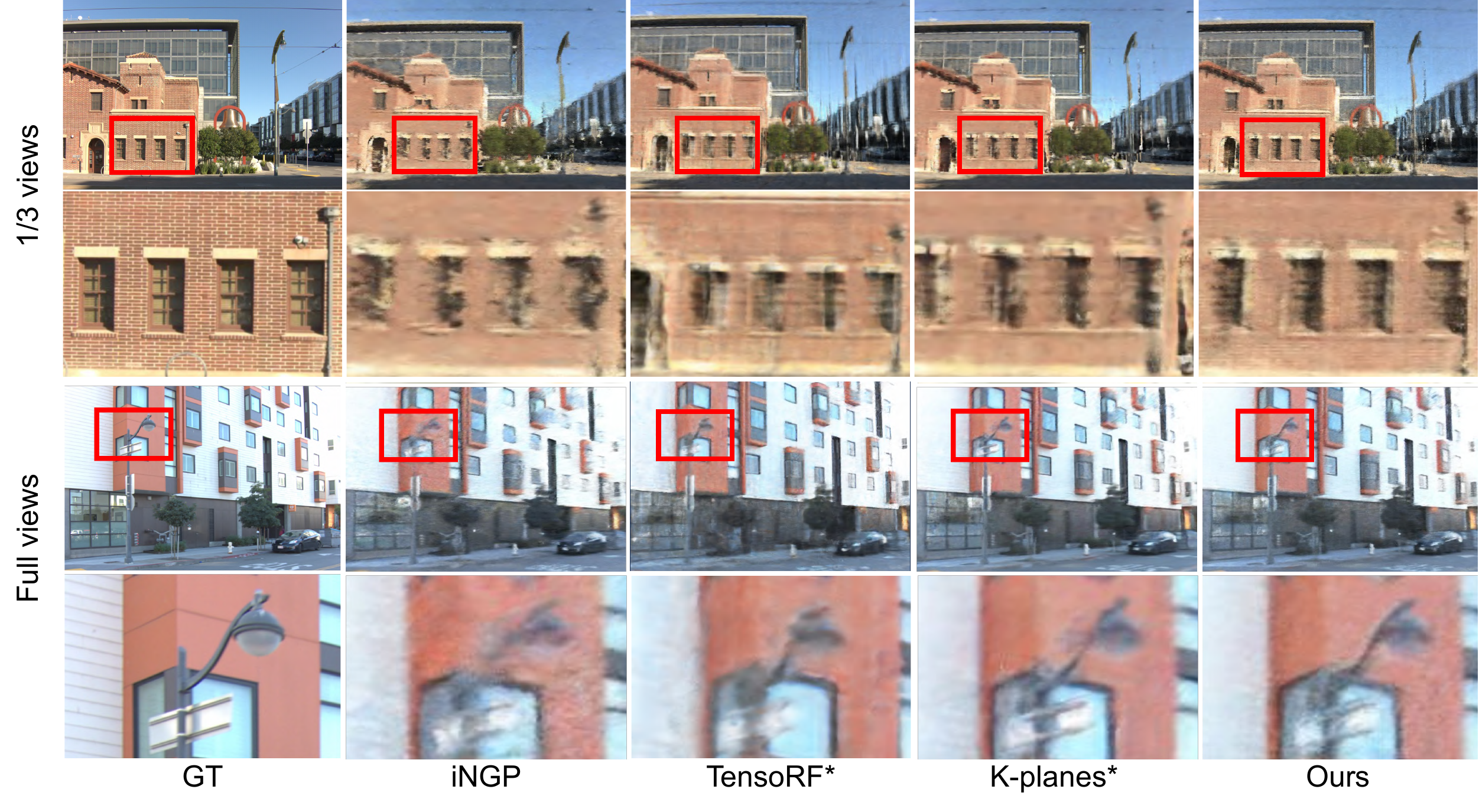}
    \vspace{-5mm}
    \caption{Qualitative results on the San Francisco Mission Bay dataset. For 1/3 views, we conduct the experiment with the default size of baselines~\cite{muller2022instant, fridovich2022plenoxels} and use 1/3 number of images from the dataset for training. For Full views, we constrain the model sizes to $\approx$ 50 Mb and train in the full-view setting. 
    }
    \vspace{-2.5mm}
    \label{fig:block_qual}
    \vspace{-2.5mm}
\end{figure}

Considering the nature of large-scale data, we opt to build the model utilizing a proposal network rather than a bitfield method as rejection sampling.
The proposal network employs an same number of samples irrespective of the encoding method.
Consequently, to assess the expressive capabilities between encoding methods, we conduct a comparison within equivalent memory constraints.
For the constrained model size setting, our model outperforms iNGP and K-planes in both the full-view and few-shot settings.
\Fref{fig:block_qual} shows that our model can represent the scene more accurately with the constrained model size.
This implies our model has superior extensibility to large-scale scene reconstruction in on-device scenarios.

\section{Additional Experiments}

\subsection{Ablation Study on Collision}
\label{sec:ablation on collision}

The Fact-Hash method imposes a constraint by confining collisions within a two-dimensional plane through a \textit{Map} process, compared against iNGP, permitting collisions across the entire 3D space. Collision rate assessment is complex due to unsampled features via bit-field sampling, leading to varying collision rates among scenes. To evaluate collision effects indirectly we use the number of model parameters in the impact of different hash table sizes.

\begin{wrapfigure}{ht}{0.20\textwidth}
    \vspace{-4mm}
    \centering
    \includegraphics[width=1\linewidth]{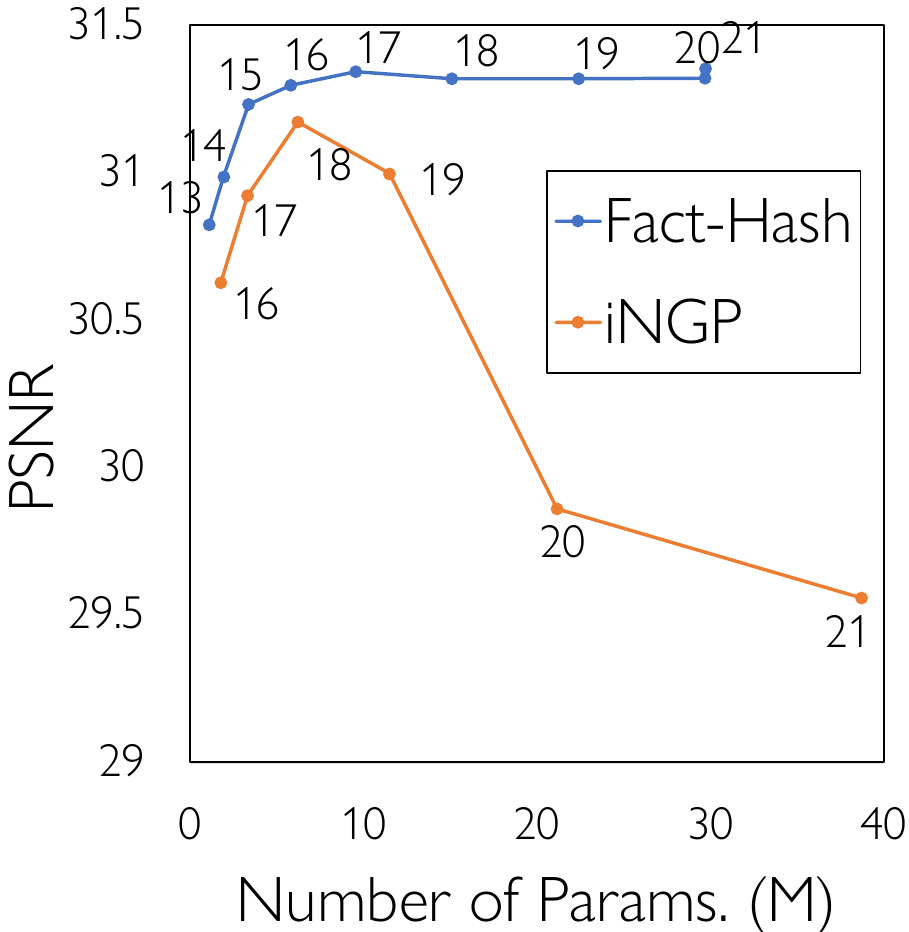}
    \caption{Comparison between iNGP and Fact-Hash in various collision rate. PSNR values are averaged along the scenes where iNGP success in training, excluding \texttt{hotdog, lego, materials}.}
    \label{fig:collision}
    \vspace{-4mm}
\end{wrapfigure}

We observe that as collision rate decreases, with higher hash table size, the PSNR values degrade (see ~\Fref{fig:collision}).
The results show instability in bitfield learning within 3D collision scenarios, which aligns with the observation explained in \jskim{\ref{sec:codebase_comparison}}.
\jskim{Higher collision rates cause learning failures, while lower collision rates degrade performance on detailed scene (\eg \texttt{ship}), both leading to PSNR decline. The use of a smaller batch size for on-device settings results in overfitting of the density distribution (\Fref{fig:bitfield}), reducing generalization capabilities.}
\jskim{However, Fact-Hash proposed 2D collision restriction reducing the probability of collision and low-rank approximation mitigating floating density artifacts. These results in superior robustness and expressiveness compared to iNGP.}

\subsection{Complementary relationship of encoding methods}
As the collision rate increases, \jskim{iNGP} shows training instability and lower PSNR compared to Fact-Hash.
We present an ablation study, extending the analysis in \jskim{~\ref{sec:ablation on collision}}. To assess the pure impact of encoding, we do not use bitfield.

\begin{wrapfigure}{ht}{0.15\textwidth}
    \centering
    \hspace{-4mm}   \includegraphics[width=1\linewidth]{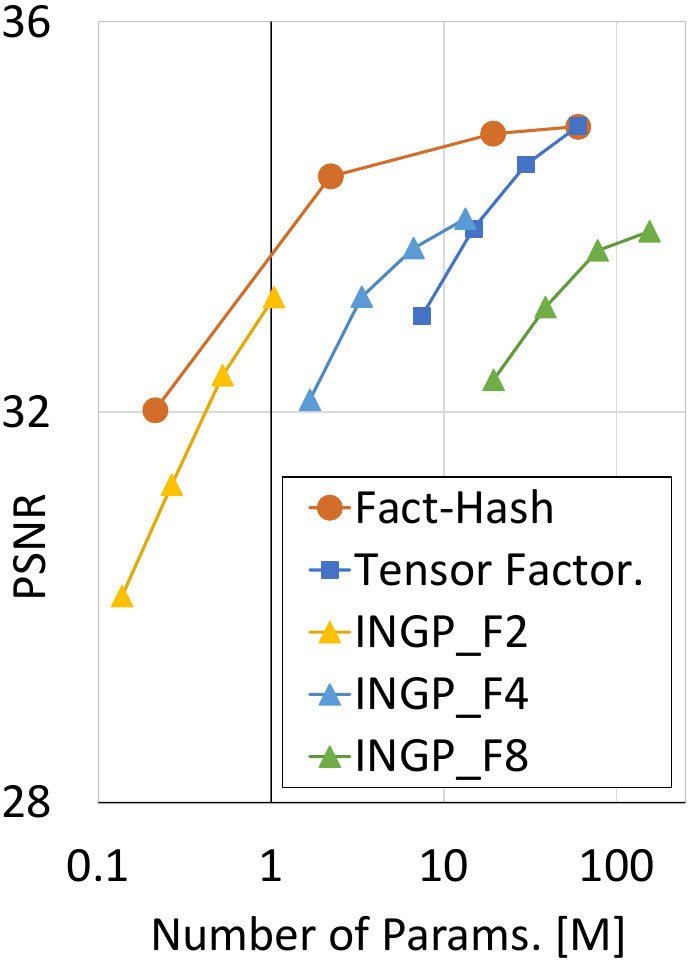}
    \vspace{-2mm}
    \caption{Ablation study on encoding methods.}
    \vspace{-4mm}
    \label{fig:analysis}
\end{wrapfigure}
To confirm synergy between Tensor Factorization and Hash-encoding, we assessed PSNR across parameters on \texttt{chair} data, by compressing Factorized encoding, by adding hash collisions (Fact-Hash) and reducing tensor rank (Tensor Factor.). \Fref{fig:analysis} shows combining Hash and Tensor Factorization exhibited improved compression performance, indicating their complementary nature in enhancing encoding efficiency.

To compare with 3D factorization, we grid-searched along the feature dim.~(1,2,4,8) and hash table size ($2^9$-$2^{21}$), but none of the configurations yielded superior performance compared to Fact-Hash.
Experiment supports that Fact-Hash leverages the complementary factors of each encoding methods, resulting in compact, expressive encoding, and robust training. This enables to extremely reduce communication costs in memory-limited on-device situations.
\section{Conclusion}
\label{sec:conclusion}
We propose Fact-Hash, a novel compact parametric encoding method for on-device execution.
Fact-hash finds the integration between 
Hash-encoding and Tensor Factorization.
Our extensive experiments 
that spans various application scenarios, including sparse inputs and real-world large-scale environments, confirms the robust operation of our method. 
We conduct on-device experiments 
under constrained power settings.
We illustrate our model, with significantly fewer parameters, achieves enhanced rendering speeds while achieving favorable 
quality of novel-view reconstructions, demonstrating the suitability of our model for on-device scenarios.
\jskim{Integration of Fact-Hash encoding to other rendering pipeline, extending its applicability~\ie diverse lighting and anti-alising, and exploration on diverse datasets to measure its scalability in larger volumes of data remains as a future works.}

{
    \small
    \bibliographystyle{IEEEtran}
    \bibliography{references}
}

\end{document}